\documentclass{article}

\usepackage[final,nonatbib]{neurips_2024_ml4ps}
\usepackage[utf8]{inputenc} % allow utf-8 input
\usepackage[T1]{fontenc}    % use 8-bit T1 fonts
\usepackage{hyperref}       % hyperlinks
\usepackage{url}            % simple URL typesetting
\usepackage{booktabs}       % professional-quality tables
\usepackage{amsfonts}       % blackboard math symbols
\usepackage{nicefrac}       % compact symbols for 1/2, etc.
\usepackage{microtype}      % microtypography
\usepackage{xcolor}         % colors
\usepackage{graphicx}
\usepackage{multicol}
\usepackage{multirow}
\usepackage{todonotes}
\usepackage{tikz}

\title{\hspace{1.5cm}RoBo6: Standardized MMT Light Curve\\\hspace{1.5cm}Dataset for Rocket Body Classification}

\author{%
  Daniel Kyselica\\
  Department of Applied Informatics\\
  Comenius University Bratislava\\
  842 48 Bratislava, Slovakia \\
  \texttt{daniel.kyselica@fmph.uniba.sk} \\
  \And
  Marek Šuppa \\
  Department of Applied Informatics\\
  Comenius University Bratislava\\
  842 48 Bratislava, Slovakia \\
  \texttt{marek.suppa@fmph.uniba.sk} \\
  \And
  Jiří Šilha \\
  Division of Astronomy and Astrophysics\\
  Comenius University Bratislava\\
  842 48 Bratislava, Slovakia \\
  \texttt{jiri.silha@fmph.uniba.sk} \\
  \And
  Roman Ďurikovič \\
  Department of Applied Informatics\\
  Comenius University Bratislava\\
  842 48 Bratislava, Slovakia \\
  \texttt{roman.durikovic@fmph.uniba.sk} \\
}

\begin{document}

    \begin{tikzpicture}[
        overlay,% Do our drawing on an overlay instead of inline
        remember picture,% Allow us to share coordinates with other drawings
        shift=(current page.north west),% Set the top (north) left (west) as the origin
        yscale=-1,% Switch the y-axis to increase down the page
    ]
        \node at (1.8in, 1.75in) {\includegraphics[height=6em]{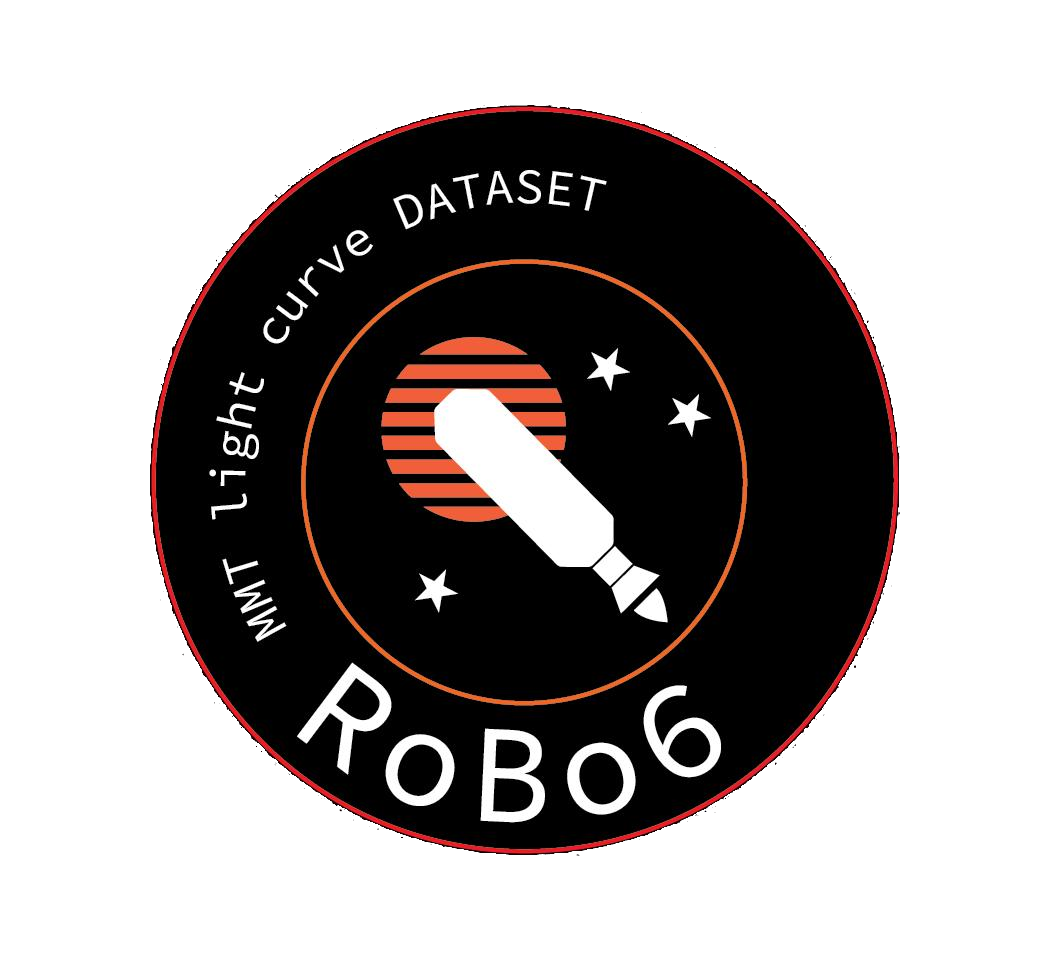}};
    \end{tikzpicture}

\maketitle

\begin{abstract}

Space debris presents a critical challenge for the sustainability of future space missions, emphasizing the need for robust and standardized identification methods. However, a comprehensive benchmark for rocket body classification remains absent. This paper addresses this gap by introducing the RoBo6 dataset for rocket body classification based on light curves. The dataset, derived from the Mini Mega Tortora database, includes light curves for six rocket body classes: CZ-3B, Atlas 5 Centaur, Falcon 9, H-2A, Ariane 5, and Delta 4. With 5,676 training and 1,404 test samples, it addresses data inconsistencies using resampling, normalization, and filtering techniques. Several machine learning models were evaluated, including CNN and transformer-based approaches, with Astroconformer reporting the best performance. The dataset establishes a common benchmark for future comparisons and advancements in rocket body classification tasks.

\end{abstract}

\section{Introduction}
%shorter version
Humans have been utilizing the space for over 60 years for various purposes, launching more than 6000 missions~\cite{ESA01}.
The consequence of this is the increasing amount of space debris, which became a serious threat to future missions~\cite{ESA03}.

Each object reflects light, creating a measurable signal called a light curve, which serves as its unique footprint. 
The light curve is shaped by the object's shape, reflectivity, geometry, surface, and rotation thus encoding the objects appearence. 
% This notion can be utilized to characterize unknown objects. 
% For example on 4 March 2022, the unknown object WE0913A impacted the far side of the Moon ~\cite{lunarImpactor}. 
% Debate over its nature pointed to two possible rocket bodies, but the object's light curve helped identify it as the Long March 3C rocket.
Although traditional methods extract limited details from light curves, some potentially crucial parameters are often overlooked.

Machine learning can help scientists identify hazardous and unknown objects more efficiently than traditional methods, 
enhancing space safety~\cite{xiang2020space}. Studies like \cite{allworth2021transfer} and \cite{yao2021basic} demonstrate its ability to distinguish between rocket bodies.
However, existing solutions are not directly comparable. Models are often evaluated solely on proprietary datasets, and even when publicly accessible datasets are used, variations in sample filtration, preprocessing, normalization, and evaluation methods can lead to substantial inconsistencies in performance assessments.
For instance, strict filtering might limit evaluation to ideal scenarios, ignoring real-world complexities. Similarly, flawed evaluation strategies might falsely favor models that excel in dominant classes, masking their overall shortcomings.

A standardized benchmark dataset is hence essential to ensure fair comparison, identify the most effective approach and establish a foundation for consistent advancements in the field.

\section{MMT Database}\label{sec:mmt}

\begin{figure}[h!]
    \centering
    \includegraphics[width=0.7\linewidth]{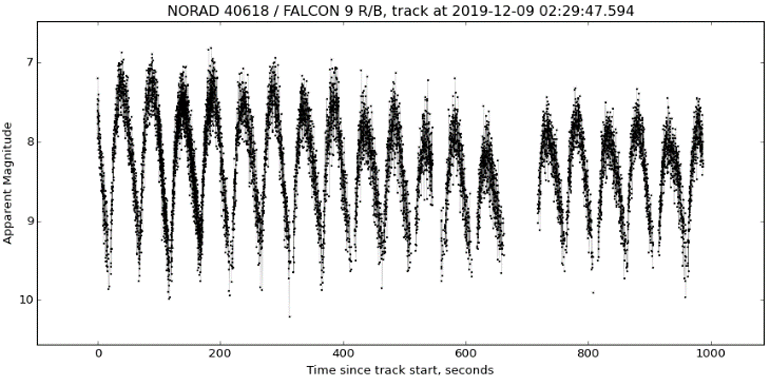}
    \caption{Example of Falcon 9 light curve from MMT database \cite{mmtWeb}.}
    \label{fig:mmt_example}
\end{figure}

The primary source of the publicly available light curves dataset is the Russian Mini Mega Tortora (MMT) database \cite{karpov2016massive,mmtWeb} a wide-field monitoring system operated by the Special Astrophysical Observatory of the Russian Academy of Sciences. 
% At the time of writing this article in August 2024,
In August 2024,
the database contained 14, 888 objects and 502, 815 tracks (or light curves),  with new records being added daily. The light curves of objects, identified by their NORAD ID, are stored in a sequence of measurements containing the time, standard magnitude, and phase angle. 
Figure \ref{fig:mmt_example} presents a sample light curve of the Falcon 9 rocket body. The database is widely used for the training and validation 
% of  machine learning models
for object characterization and attitude determination, described in Section~\ref{sec:mmt_research}.

The data, however, exhibits several significant flaws: large gaps between observations, a limited number of observations per rotational period, and a low signal-to-noise ratio. These imperfections arise from physical processes, varying observation conditions, the high variability of the signal, and the inherent characteristics of the sensors. It is hence imperative for the raw data to be pre-processed before it can be used to train Machine learning models, as discussed in Section \ref{sec:dataset}.

\section{Machine learning research using MMT data}\label{sec:mmt_research}

In recent years, there has been a trend to extract information about space objects from their light curves. A common kind of information to extract is the shape of an object. In this work, we focus on the shape classification problem with MMT be the main data source.
% Machine learning models need data to learn from, making the MMT database a popular choice thanks to its size and availability.

One of the first utilization of the database for machine learning was by the authors in \cite{furfaro2018space}. They have set up a classification task using a custom 1-D convolutional network with 3 different target classes, namely: rocket bodies, debris, and satellites. It uses a sequence of 500 measurements from the light curve as input, with the reported test accuracy of around 75\%. 

A similar approach was used in \cite{linares2020space} with the same input sequence strategy and 1-D CNN achieved a comparable accuracy of 75\% with a different target class. To preserve more information in the input, in \cite{allworth2021transfer} the authors used phase angle data alongside magnitude measurements, resulting in two channel sequences of 1200 points in length, covering around 20 minutes of observation time. To achieve uniform size between light curves, truncation and zero padding was used. This helped the 1-D CNN network to distinguish between 8 classes (7 rocket body classes + bow-wing class) with a 90\% accuracy. To enhance the performance, the model was pre-trained using simulated data.
To ensure uniform temporal sampling, in the work \cite{kerr2021light} the inputs were resampled to a consistent frequency and a 1-D CNN and LSTM-based models were trained to classify object shapes specifically within the GEO orbit.
As in the previous work, the data were padded and truncated to a uniform size. Normalization was done by rescaling the sample between 0 and 1 by its maximum and minimum values.

% To ensure uniform temporal sampling of the input sequence, the authors in \cite{kerr2021light} resampled all light curves to a consistent frequency. Then they trained their 1-D CNN and LSTM-based models to classify object shapes specifically within the GEO orbit. As in the previous work, the data were padded and truncated to a uniform size. Normalization was done by rescaling the sample between 0 and 1 by its maximum and minimum values.

In \cite{yao2021basic}, a different approach was adopted to standardize input size while minimizing information loss. Light curves were folded by their rotational period, producing a new sample with 200 measurements each. 
This re-processing technique, combined with data augmentation approaches such as  phase translation and noise addition, enabled a 1-D CNN to achieve 85\% accuracy in a classification scenario involving 5 target classes of rocket bodies. Building on prior work, the authors in \cite{kyselica2022astronomical} implemented a method which involved segmenting light curves into individual rotational periods, and rescaling them into uniform size, resulting in a significantly expanded training dataset. Using this approach, a  1-D version of ResNet20~\cite{he2016deep} network was employed to classify three rocket body classes, achieving 84\% accuracy. 

A contemporary approach to handling large sequences is using the Transformer\cite{vaswani2017attention} module. Astroconformer, a Transformer-based model introduced \cite{shrive2024classifying}, was originally designed for 
surface gravity estimation from stellar light curves processing 
% stellar light curves, with its primary application being the estimation of surface gravity. 
,yet it can be easily adapted for a classification tasks.
 
While this section highlights only a subset of works in the subfield, it is evident that substantial differences exist in the preprocessing, evaluation, and normalization strategies employed across studies. These inconsistencies make direct comparison between methods highly challenging, if not impossible, emphasizing the need for a standardized approach to enable meaningful evaluation. 

\section{RoBo6 dataset}\label{sec:dataset}

Prior work has predominantly focused on classifying objects such as rocket bodies, debris, or satellites based on their light curves. However, a more practically relevant challenge lies in distinguishing between objects with a similar shape, such as different types of rocket bodies. To the best of our knowledge, no standard benchmark currently exists for addressing this specific classification problem. 

To address this gap, we curated a standardized dataset comprising six common rocket body populations 
% with sufficient data
: CZ-3B, Atlas 5 Centaur, Falcon 9, H-2A, Ariane 5 and Delta 4. 
The dataset is divided into a train set of 5,676 samples and a test set of 1,404 samples, with further details provided in Table \ref{tab:dataset1}.

\begin{table}[h]
    \centering
    \caption{Number of 
samples per split and class. }
    \label{tab:dataset1}
    \begin{tabular}{c|c|c|c|c|c|c}
\toprule
& \textbf{Ariane 5} & \textbf{Delta 4} & \textbf{CZ-3B} & \textbf{Atlas 5 Centaur} & \textbf{H-2A} & \textbf{Falcon 9}\\
\midrule
\textbf{Train} & 660 & 233 & 2266 & 1029 & 623 & 865 \\
\textbf{Test} & 173 & 70 & 548 & 247 & 150 & 216 \\
\bottomrule
    \end{tabular}

\end{table}

% Each sample is described by the following fields:
% \begin{itemize}
%     \item Label - the name of the rocket body population;
%     \item Id - unique identification number of the original light curve;
%     \item Part - Serial number of the sample after splitting of the source light curve;
%     \item Period - Apparent rotational period in 0.1 seconds;
%     \item Mag - Path to file containing 10000 standard magnitude values;
%     \item Phase - Path to file containing 10000 phase angle values;
%     \item Time - Path to file containing 10000 timestamp values in nanoseconds;
% \end{itemize}
Each sample is characterized by the following fields: \textit{label}, \textit{ID}, \textit{part} number, \textit{period} (in nanoseconds), \textit{mag}, \textit{phase} and \textit{time}. The last three fields refer to file paths that contain additional metadata, such as standard magnitude, phase angle, and time measurements. 
During preprocessing, samples can be divided into subsamples, with their sequence order recorded in the \textit{part} field.

Since many machine learning models, such as CNNs, require grid structured data, each sample needs to be resampled onto a uniformly spaced grid with a manageable length. In order to convert the data into this format and to retrain as much information as possible, a series of filtering, splitting and resampling operations was performed over each sample. Standard normalization using mean and standard deviation was also applied.

The dataset is publicly available on the Hugging Face platform  (\url{https://huggingface.co/datasets/kyselica/RoBo6}) and was created from data downloaded in August 2024 from the official MMT website~\cite{mmtWeb} using a custom Python script.

\subsection{Gap-Based Splitting and Frequency Rescaling}

The original dataset contains light curves that often span extended periods. However, in many cases, the gap between two consecutive measurements is larger than the rotational period of the object, presenting an opportunity to split such samples at these gaps. This process results in shorter light curves, as the intermediate gaps do not contain meaningful information. 

Following this gap-based splitting, 95\% of the samples were found to be shorter than 1,000 seconds. For the remaining longer samples, additional splits were made at this 1,000-second threshold. A sampling frequency of 10 Hz was chosen based on an analysis of time intervals between consecutive measurements. To standardize the dataset, all samples were rescaled and zero-padded to a fixed length of 10,000 points, representing 1,000 seconds of data.

% To create a grid a sampling frequency needs to be chosen with respect to the original data. By analysis of the gaps between two consecutive measurements, we arrive to the conclusion that the base frequency is around $0.1$. Using this value, samples are rescaled and zero-padded to 10 000 points corresponding to 1000 seconds worth of data.

% \subsection{Spliting}

% Original data contains light curves spanning over a very long time. However, in many cases, the gap between two consecutive measurements is larger than the rotational period of the object. This gives an opportunity to split such samples by the gap resulting in two shorter light curves as there is no useful information in this period. 
% % The length statistics before and after splitting are presented in Tab.\ref{tab:split}.
% 1000 seconds was decided to be the sample's length for the dataset, as 95\% of the samples have smaller lengths after splitting. Longer samples are simply split into halves until the size is under the selected value.

% \begin{table}[h]
%     \centering
%     \caption{Light curves' length in 0.1 sec. before and after splitting by gaps.}
%     \label{tab:split}
%     \begin{tabular}{|c|c|c|c|c|c|}
%     \cline{2-6}
%         \multicolumn{1}{c|}{} & \textbf{Max} & \textbf{Min} & \textbf{Mean} & \textbf{Median} & \textbf{Std}  \\ \hline
%      \textbf{Original}   & 301001 & 1 & 12017.46 & 7157  & 15936.59 \\ \hline
%     \textbf{After split} & 62539 & 1 & 3384.18 & 1909 & 4042.38 \\
%     \hline
%     \end{tabular}

% \end{table}

\subsection{Filtering of Low-Quality Samples}

Low-quality samples are filtered based on two criteria emprically determined citeria consistent with values commonly used in prior studies, such as \cite{furfaro2018space}. 

The first criterion sets the minimum number of measurements in one sample to 100. The second criterion evaluates the coverage of the apparent rotational period using a folded light curve, which is created by folding the data based on the object's apparent rotational period \cite{vsilha2021light} and reshaped to 100 points. To ensure sufficient data quality, the folded light curve must have at least 75\% coverage by measurements.

\subsection{Evaluation Strategy and Metric Selection}

Selecting an effective evaluation method is essential for the precise assessment of model performance. A viable strategy involves partitioning data by objects into train and test sets to gauge the model's competence in classifying new objects. Nevertheless, this method is challenging with multiple classes due to the limited object count and a significantly imbalanced light curve distribution, potentially resulting in an inadequately representative test set and insufficient samples for reliable evaluation. As shown in Table \ref{tab:dataset2}, the disparity in data distribution is particularly evident in the case of the CZ-3B rocket, where just 14 objects account for more than $60\%$ of all light curve tracks.

\begin{table}[h]
    \centering
    \caption{Number of tracks per CZ-3B rocket.}
    \label{tab:dataset2}
    \begin{tabular}{c|c|c|c}
\toprule
\textbf{\#Track Range} & \textbf{N.o. objects} & \textbf{N.o. tacks} & \textbf{Dataset Coverage} \\
\midrule
326 - 391& 1 & 391 & 7.87 \% \\
261 - 326& 2 & 546 & 10.99 \% \\
196 - 261& 6 & 1303 & 26.22 \% \\
131 - 196& 5 & 782 & 15.74 \% \\
66 - 131 & 10 & 907 & 18.25 \% \\
1 - 66   & 63 & 1040 & 20.93 \% \\
\bottomrule
    \end{tabular}

\end{table}

In real-world scenarios, the same object can be observed multiple times and must be consistently identified. To simulate this, the dataset can be split by track ID, ensuring all samples derived from the same light curve during preprocessing are assigned to the same set. Stratified splitting is employed to maintain consistent class distributions across the training and testing sets.

Given the imbalanced nature of the dataset, accuracy alone can be a misleading metric, since strong performance on classes with many samples may overshadow poor performance on underrepresented classes.
To address this issue, the F1 macro score was chosen as the primary evaluation metric, as it takes into account the cardinality of all classes and thus provides a more balanced assessment of model performance across the dataset.

\subsection{Model Training and Performance Evaluation}

To demonstrate the utility of the dataset, five selected models \cite{allworth2021transfer},~\cite{he2016deep},~\cite{furfaro2018space},~\cite{yao2021basic} and \cite{pan2024astroconformer} were trained on it. Each model was trained for 50 epochs using the Adam optimizer with a learning rate of $0.001$. Training parameters specified in the respective publications were followed; in cases where parameters were not provided, default settings were used. The complete list of parameters and further preprocessing details can be found in  Table \ref{tab:params}.
The results, summarized in Table \ref{tab:results}, align closely with those reported in the respective publications. Astroconformer, despite being developed for stellar light curves, proved to be the best-performing model in this benchmark comparison.

\begin{table}[ht]
    \centering
    \caption{Models performance.}
    \label{tab:results}
    \begin{tabular}{c|c|c|c|c}
\toprule
\textbf{Model} & \textbf{Accuracy} & \textbf{F1} & \textbf{Precision} & \textbf{Recall} \\
\midrule
ALLWORTH~\cite{allworth2021transfer}             & 0.559 ± 0.044 & 0.478 ± 0.038 & 0.491 ± 0.033 & 0.531 ± 0.024 \\
RESNET~\cite{he2016deep}               & 0.694 ± 0.023 & 0.600 ± 0.034 & \textbf{0.738} ± 0.026 & 0.584 ± 0.033 \\
FURFARO~\cite{furfaro2018space}              & 0.628 ± \textbf{0.009} & 0.552 ± \textbf{0.013} & 0.570 ± 0.017 & 0.552 ± \textbf{0.013} \\
YAO~\cite{yao2021basic}                  & 0.672 ± 0.017 & 0.604 ± 0.023 & 0.622 ± 0.029 & 0.601 ± 0.020 \\
ASTROCONFORMER~\cite{pan2024astroconformer}       & \textbf{0.725} ± 0.011 & \textbf{0.684} ± 0.015 & 0.702 ± \textbf{0.010} & \textbf{0.677} ± 0.019 \\
\bottomrule
    \end{tabular}

\end{table}

The hyperparameters used for the training the respective models are detailed in  Table~\ref{tab:params}.  In an attempt to reproduce the published models as closely to their published state as possible, a specific preprocessing step was incorporated specifically for these models:

\begin{itemize}
    \item \textbf{ALLWORTH}~\cite{allworth2021transfer} Observations from the initial 20 minutes were rescaled to a uniform grid consisting of 1,200 points.
    \item \textbf{FURFARO}~\cite{furfaro2018space} - The first 500 points were utilized as input.
    \item \textbf{YAO}~\cite{yao2021basic} - The light curve was folded with the apparent rotation period to 200 point grid.
\end{itemize}

\begin{table}[h!]
    \centering
    \caption{Parameters used in the training process for each model.}
    \label{tab:params}
    \begin{tabular}{c|c|c|c|c}
    \toprule
         \multicolumn{1}{c}{} & \textbf{Input size} & \textbf{Channels} &  \textbf{Batch size} & \textbf{Scheduler} \\ \midrule
         \textbf{Default} & 10 000 & Mag &  32 &  x \\ 
         \textbf{ALLWORTH\cite{allworth2021transfer}} & 1 200 & Mag + Phase & 256 & x \\ 
         \textbf{RESNET\cite{he2016deep}} & 10 000 & Mag &  32 & x \\ 
         \textbf{FURFARO\cite{furfaro2018space} } & 500 & Mag &  32 & x \\ 
         \textbf{YAO\cite{yao2021basic}  } & 200 & Mag &  32 & x \\ 
         \textbf{ASTROCONFORMER\cite{pan2024astroconformer}} & 10 000 & Mag  & 32 & \checkmark \\ 
         \bottomrule

    \end{tabular}
\end{table}

Only one paper offers a basis for comparison. We trained the Allworth method~\cite{allworth2021transfer} on our dataset and compared the results with those reported in the origianl publication. However, the comparison is not entirely fair, as the training scenarios differ. Specifically, the authors in~\cite{allworth2021transfer} used a balanced dataset with only 500 samples per class. Table \ref{tab:allworth} shows the result for the overlapping  classes between the two datasets. The source code for the experiments is publickly available at \url{https://github.com/kyselica12/RoBo6_Model_Comparison}.

\begin{table}[h!]
    \centering
    \caption{Comparison of the Allworth method\cite{allworth2021transfer} on our dataset and the original paper.}
    \label{tab:allworth}
    \begin{tabular}{c|c|c|c|c|c|c}
\toprule
\multicolumn{1}{c|}{\textit{\textbf{Rocket Bodies}}} & \multicolumn{2}{|c|}{\textbf{F1 score}} & \multicolumn{2}{|c|}{\textbf{Precision}} & \multicolumn{2}{|c|}{\textbf{Recall}} \\
 \multicolumn{1}{c|}{} & \textbf{Our} & \textbf{Original}  & \textbf{Our} & \textbf{Original}  & \textbf{Our} & \textbf{Original} \\ 
\midrule
Atlas 5 Centaur & 0.39 & 0.74 & 0.53 & 0.71 & 0.33 & 0.78 \\
CZ-3B & 0.65 & 0.63 & 0.77 & 0.61 & 0.57 &  0.65 \\
Delta 4 & 0.11 & 0.81 & 0.19 & 0.78 & 0.09 & 0.85  \\
Falcon 9 & 0.49 & 0.80 & 0.48 & 0.81 & 0.49 & 0.79 \\
H-2A & 0.56 & 0.79 & 0.46 & 0.82 & 0.72 & 0.77 \\
\bottomrule
    \end{tabular}
    
\end{table}

\section{Conclusion}

We introduce the RoBo6 dataset, a new benchmark for rocket body classification based on light curves. Derived from the MMT database, it includes six classes of rocket bodies and addresses inconsistencies in prior datasets through preprocessing and standardization. By enabling easy comparison across models and providing robust training and testing splits, RoBo6 facilitates consistent evaluations and accelerates advancements in this research area. Models trained on the dataset demonstrated its suitability for both traditional and transformer-based architectures, with Astroconformer achieving the highest performance. We hope this dataset fosters advancements in space object classification and promotes sustainable practices in space exploration.

\medskip

\bibliographystyle{unsrt}
\bibliography{references}

\appendix

\end{document}